\documentclass{article}

\usepackage{microtype}
\usepackage{graphicx}
\usepackage{booktabs} 

\usepackage{enumitem}
\usepackage{xspace}
\usepackage{subcaption}
\usepackage{amsmath,amsthm,amssymb,amsfonts}
\usepackage{makecell}
\usepackage{array}
\usepackage[firstpage]{draftwatermark}
\SetWatermarkText{
 \hspace*{2.25in}
 \raisebox{10in}{
  \includegraphics[height=0.9in]{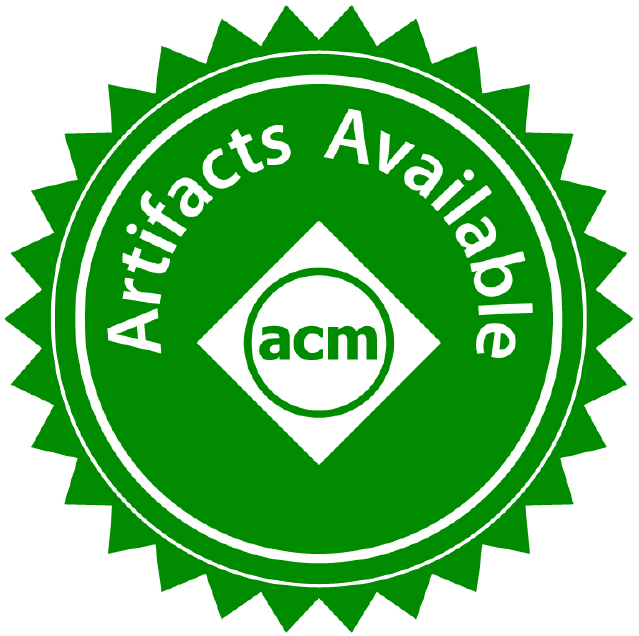}
  \includegraphics[height=0.9in]{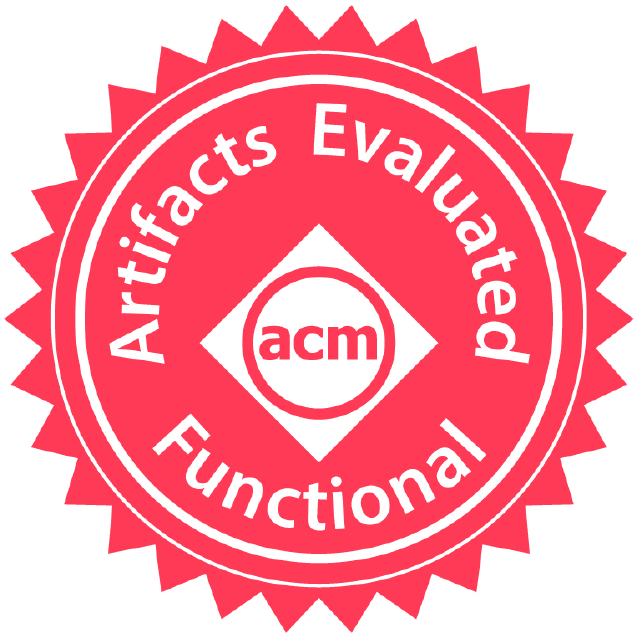}
  \includegraphics[height=0.9in]{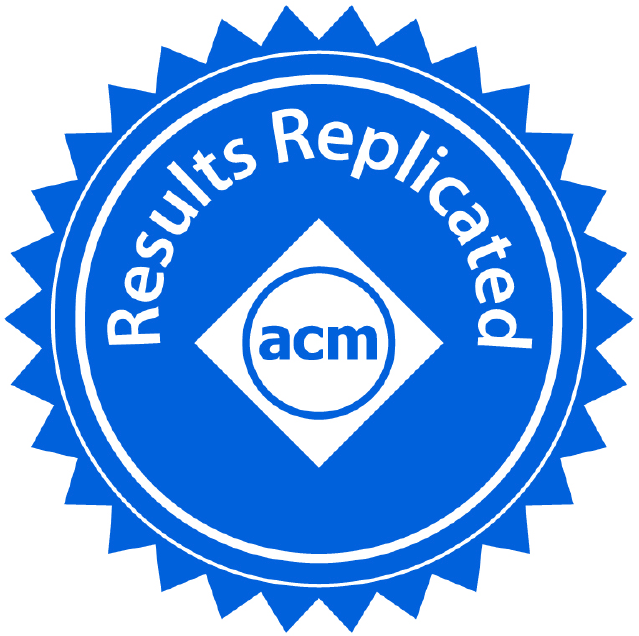}
 }
}
\SetWatermarkAngle{0}

\usepackage{hyperref}

\usepackage[accepted]{mlsys2020}

\newcolumntype{L}[1]{>{\centering\arraybackslash}m{#1}}
\newcommand{\minihead}[1]{{\vspace{.25em}\noindent\textbf{#1}\xspace}}

\newcommand{\mlperf}{MLPerf\xspace}

\mlsystitlerunning{MLPerf Training Benchmark}

\begin{document}

\twocolumn[
\mlsystitle{MLPerf Training Benchmark}




\begin{mlsysauthorlist}
\mlsysauthor{Peter Mattson}{go}
\mlsysauthor{Christine Cheng}{intel}
\mlsysauthor{Cody Coleman}{stanford}
\mlsysauthor{Greg Diamos}{landing}
\mlsysauthor{Paulius Micikevicius}{nvidia}
\mlsysauthor{David Patterson}{go,berkeley}
\mlsysauthor{Hanlin Tang}{intel}
\mlsysauthor{Gu-Yeon Wei}{harvard}
\mlsysauthor{Peter Bailis}{stanford}
\mlsysauthor{Victor Bittorf}{go}
\mlsysauthor{David Brooks}{harvard}
\mlsysauthor{Dehao Chen}{go}
\mlsysauthor{Debojyoti Dutta}{cisco}
\mlsysauthor{Udit Gupta}{harvard}
\mlsysauthor{Kim Hazelwood}{fb}
\mlsysauthor{Andrew Hock}{cerebras}
\mlsysauthor{Xinyuan Huang}{cisco}
\mlsysauthor{Atsushi Ike}{fujitsu}
\mlsysauthor{Bill Jia}{fb}
\mlsysauthor{Daniel Kang}{stanford}
\mlsysauthor{David Kanter}{rwt}
\mlsysauthor{Naveen Kumar}{go}
\mlsysauthor{Jeffery Liao}{synopsys}
\mlsysauthor{Guokai Ma}{intel}
\mlsysauthor{Deepak Narayanan}{stanford}
\mlsysauthor{Tayo Oguntebi}{go}
\mlsysauthor{Gennady Pekhimenko}{toronto,vector}
\mlsysauthor{Lillian Pentecost}{harvard}
\mlsysauthor{Vijay Janapa Reddi}{harvard}
\mlsysauthor{Taylor Robie}{go}
\mlsysauthor{Tom St. John}{tesla}
\mlsysauthor{Tsuguchika Tabaru}{fujitsu}
\mlsysauthor{Carole-Jean Wu}{fb}
\mlsysauthor{Lingjie Xu}{alibaba}
\mlsysauthor{Masafumi Yamazaki}{fujitsu}
\mlsysauthor{Cliff Young}{go}
\mlsysauthor{Matei Zaharia}{stanford}
\end{mlsysauthorlist}

\mlsysaffiliation{go}{Google}
\mlsysaffiliation{intel}{Intel}
\mlsysaffiliation{stanford}{Stanford University}
\mlsysaffiliation{landing}{Landing AI}
\mlsysaffiliation{nvidia}{NVIDIA}
\mlsysaffiliation{berkeley}{University of California, Berkeley}
\mlsysaffiliation{harvard}{Harvard University}
\mlsysaffiliation{cisco}{Cisco}
\mlsysaffiliation{fb}{Facebook}
\mlsysaffiliation{cerebras}{Cerebras}
\mlsysaffiliation{rwt}{Real World Technologies}
\mlsysaffiliation{synopsys}{Synopsys}
\mlsysaffiliation{toronto}{University of Toronto}
\mlsysaffiliation{tesla}{Tesla}
\mlsysaffiliation{alibaba}{Alibaba}
\mlsysaffiliation{fujitsu}{Fujitsu}
\mlsysaffiliation{vector}{Vector Institute}

\mlsyscorrespondingauthor{Peter Mattson}{petermattson@google.com}

\mlsyskeywords{Machine Learning, SysML}

\vskip 0.3in

\begin{abstract}
Machine learning (ML) needs industry-standard performance benchmarks to support design and competitive evaluation of the many emerging software and hardware solutions for ML. But ML training presents three unique benchmarking challenges absent from other domains: optimizations that improve training throughput can increase the time to solution, training is stochastic and time to solution exhibits high variance, and software and hardware systems are so diverse that fair benchmarking with the same binary, code, and even hyperparameters is difficult. We therefore present MLPerf, an ML benchmark that overcomes these challenges. Our analysis quantitatively evaluates MLPerf's efficacy at driving performance and scalability improvements across two rounds of results from multiple vendors.
\end{abstract}
]



\printAffiliationsAndNotice{}  

\section{Introduction}
\label{sec:intro}

Machine learning (ML) has revolutionized numerous domains, including computer vision~\cite{krizhevsky2012imagenet}, language processing~\cite{devlin2018bert, radford2019language}, speech recognition~\cite{hinton2012deep}, and gaming~\cite{silver2018general, mnih2013playing, chan_2018}. Much of this progress owes to deep learning (DL), which involves training of large deep-neural-network (DNN) models on massive data sets. To keep up with this growing computational demand, hardware and software systems have garnered sizable investments~\cite{amodei2018ai}.

As the number of hardware and software systems for DL training increases~\cite{paszke2017pytorch, abadi2016tensorflow, chen2015mxnet, jia2014caffe, jouppi2017datacenter, chen2018tvm, markidis2018nvidia, intel2019bigdl}, so does the need for a comprehensive benchmark. History shows that benchmarks accelerate progress~\cite{hennessy2011computer}; for example, breakthroughs in microprocessor and relational-database systems in the 1980s inspired industry consortiums to create \textit{Standard Performance Evaluation Corporation} (SPEC) for Unix servers~\cite{dixit1991spec} and the \textit{Transaction Processing Performance Council} (TPC) for transaction processing and databases~\cite{council2005transaction}. These organizations helped develop and maintain benchmarks that their respective communities then embraced. Their success inspired the formation of MLPerf, a consortium of commercial and academic organizations, to design a comprehensive benchmark suite for DL. 

Unlike other computational workloads, DL allows a range of statistical, hardware, and software optimizations that can change the mathematical semantics of the underlying operators. Although these optimizations can boost \textit{performance} (i.e., training speed), some change the learning dynamics and affect the final model's \textit{quality} (i.e., accuracy). Even accommodating different system scales (e.g., varying the number of chips) requires changing hyperparameters, potentially affecting the amount of computation necessary to reach a particular quality target. By contrast, other compute benchmarks can evaluate systems through targeted microbenchmarks.

DL is also intrinsically approximate and stochastic, allowing multiple equally correct solutions---unlike conventional computing, which tends to allow just one correct solution. As a result, implementations and training times can vary while the final quality remains the same. Since it is approximate, DL requires careful definition of equally valid solution classes and the appropriate degrees of freedom.

Prior work has varied in granularity but has either left the above challenges unaddressed or lacked critical workloads representative of modern ML. Microbenchmarks such as DeepBench~\cite{baidu2017deepbench} are affordable to run and enable a fair comparison of competing systems by isolating hardware and software from statistical optimizations, but they fail to reflect the complexity of real workloads and have limited utility. Although throughput benchmarks like Fathom and TBD~\cite{, adolf2016fathom, zhu2018tbd, google2017benchmarks} evaluate full model architectures across a broad range of tasks to better reflect the diversity and complexity of real workloads, they limit model architecture and training innovations that advance the state-of-the-art. DAWNBench~\cite{coleman2017dawnbench} measures end-to-end training time, subject to a quality threshold (i.e., time to train), and it accommodates innovative solutions (i.e., new model architectures and training techniques, such as progressive resizing and cyclic learning rates). It additionally collects source code to promote reproducibility. DAWNBench's flexibility, however, also made it difficult to draw fair comparisons between hardware and software platforms. \mlperf builds on the strengths of prior work; it combines a broad set of benchmarks like Fathom or TBD, an end-to-end training metric like DAWNBench, and the backing of a broad consortium like SPEC.

MLPerf aims to create a representative benchmark suite for ML that fairly evaluates system performance to meet five high-level goals:
\begin{itemize}[topsep=0pt]
\item Enable fair comparison of competing systems while still encouraging ML innovation.
\item Accelerate ML progress through fair and useful measurement.
\item Enforce reproducibility to ensure reliable results.
\item Serve both the commercial and research communities.
\item Keep benchmarking effort affordable so all can participate.
\end{itemize}

This paper focuses on the design and rationale for the \mlperf Training benchmark (a related \mlperf Inference benchmark is beyond the present scope). Although prior ML benchmarking efforts~\cite{coleman2017dawnbench, adolf2016fathom, google2017benchmarks, baidu2017deepbench, zhu2018tbd} each contributed to meeting one or more of the above goals, we created MLPerf to address \textit{all} of them holistically, building on the lessons learned from these efforts. To this end, \mlperf Training does the following: 

\begin{itemize}[topsep=0pt]
\item Establish a comprehensive benchmark suite that covers diverse applications, DNN models, and optimizers.
\item Create reference implementations of each benchmark to precisely define models and training procedures.
\item Establish rules that ensure submissions are equivalent to these reference implementations and use equivalent hyperparameters.
\item Establish timing rules to minimize the effects of stochasticity when comparing results.
\item Make submission code open source so that the ML and systems communities can study and replicate the results.
\item Form working groups to keep the benchmark suite up to date.
\end{itemize}

The rest of the paper is organized as follows. In \S~\ref{sec:background}, we discuss the main challenges to benchmarks for DL training, as well as related prior work. In \S~\ref{sec:benchmark}, we review the benchmarks in our suite, the time-to-train metric, and quality thresholds. In \S~\ref{sec:process}, we describe the submission, review, and reporting of results for the various categories. Finally, in \S~\ref{sec:results} and \S~\ref{sec:discussion}, we review progress between the first two \mlperf benchmarking rounds, along with future work directions.

\section{Background}

\label{sec:background}

We begin by describing in \S~\ref{sec:challenges} the unique challenges of benchmarking ML relative to other compute tasks \cite{Dongarra1988linpack, council2005transaction} and then review prior ML-benchmarking efforts in \S~\ref{sec:prior_work}.

\subsection{Unique Challenges of Benchmark Training}
\label{sec:challenges}
ML benchmarking faces unique challenges relative to other compute benchmarks, such as LINPACK~\cite{Dongarra1988linpack} and SPEC~\cite{dixit1991spec}, that necessitate an end-to-end approach. After an ML practitioner selects a data set, optimizer, and DNN model, the system trains the model to its state-of-the-art \textit{quality} (e.g., Top-1 accuracy for image classification). Provided the system meets this requirement, the practitioner can make different operation, implementation, and numerical-representation choices to maximize system \textit{performance}---that is, how fast the training executes. Thus, an ML performance benchmark must ensure that systems under test achieve state-of-the-art quality while providing sufficient flexibility to accommodate different implementations. This tradeoff between quality and performance is challenging because multiple factors affect both the final quality and the time to achieve it.

\subsubsection{Effect of Optimizations on Quality}
\label{sec:optimization_challenge}
Although many optimizations immediately improve traditional performance metrics such as throughput, some can decrease the final model quality, an effect that is only observable by running an entire training session. For example, the accuracy difference between single-precision training and lower-precision training only emerges in later epochs~\cite{zhu2016trained}. Across several representation and training choices, the validation-error curves may only separate after tens of epochs, and some numerical representations never match the final validation error of full-precision training (lower validation error directly corresponds to higher accuracy: $\text{accuracy} = 1 - \text{error}_{\text{validation}}$). Thus, even though microbenchmarks~\cite{baidu2017deepbench, chetlur2014cudnn} can assess an optimization's performance impact, a complete training session is necessary to determine the quality impact and whether the model achieves the desired accuracy. Owing to the introduction of systems with varying numerics ~\cite{abadi2016tensorflow,banner2018scalable,kster2017flexpoint,micikevicius2017mixed} and performance optimizations, ML benchmarks must include accuracy metrics.


\subsubsection{Effect of Scale on Time to Train}
\label{sec:scale_challenge}
ML training on large distributed systems with many processors typically involves data parallelism and large minibatches to maximize system utilization and minimize training time. In turn, these large minibatches require adjustments to optimizer parameters, such as the learning rate~\cite{Krizhevsky14,goyal2017accurate}. Together, these changes affect the learning dynamics and can alter the number of iterations required to achieve the target accuracy. For example, MLPerf v0.5 ResNet-50 takes about 64 epochs to reach the target Top-1 accuracy of 74.9\% at a minibatch size of 4K,~\footnote{Source: \mlperf v0.5 results (\url{https://mlperf.org/training-results-0-5}).} whereas a minibatch size of 16K can require more than 80 epochs to reach the same accuracy, increasing computation by 30\%. Larger minibatches, however, permit efficient scaling to larger distributed systems, reducing the time to train the model. The tradeoffs between system size, minibatch size, and learning dynamics present another challenge for a DL-focused performance benchmark.

\subsubsection{Run-to-Run Variation}
\label{sec:variation_challenge}
\begin{figure}[t]
  \centering
  \begin{subfigure}{\columnwidth}
  \centering
  \includegraphics[width=0.85\columnwidth]{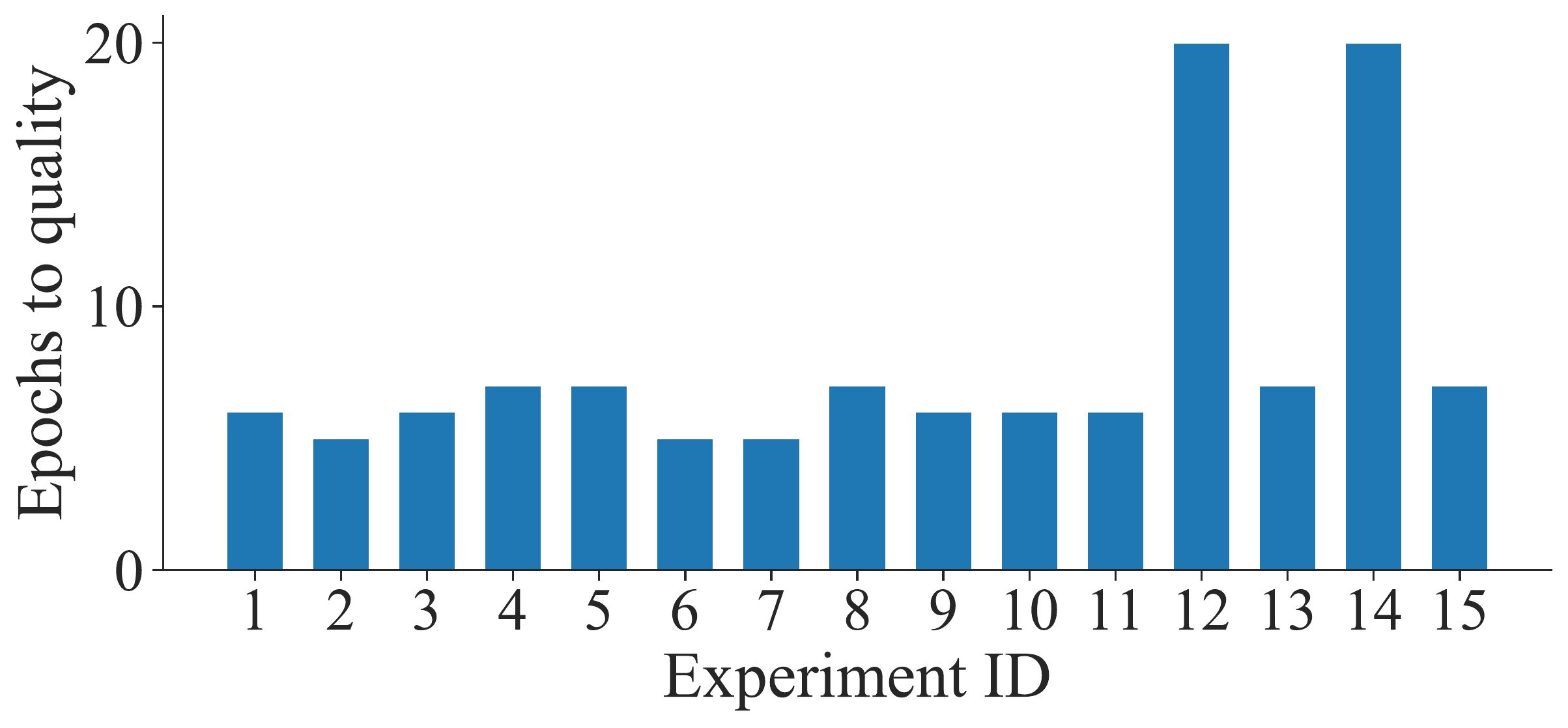}
  \caption{NCF.}
  \label{fig:epoch_variation_ncf}
  \end{subfigure}\vspace{1mm}
  
  \begin{subfigure}{\columnwidth}
  \centering
  \includegraphics[width=0.85\columnwidth]{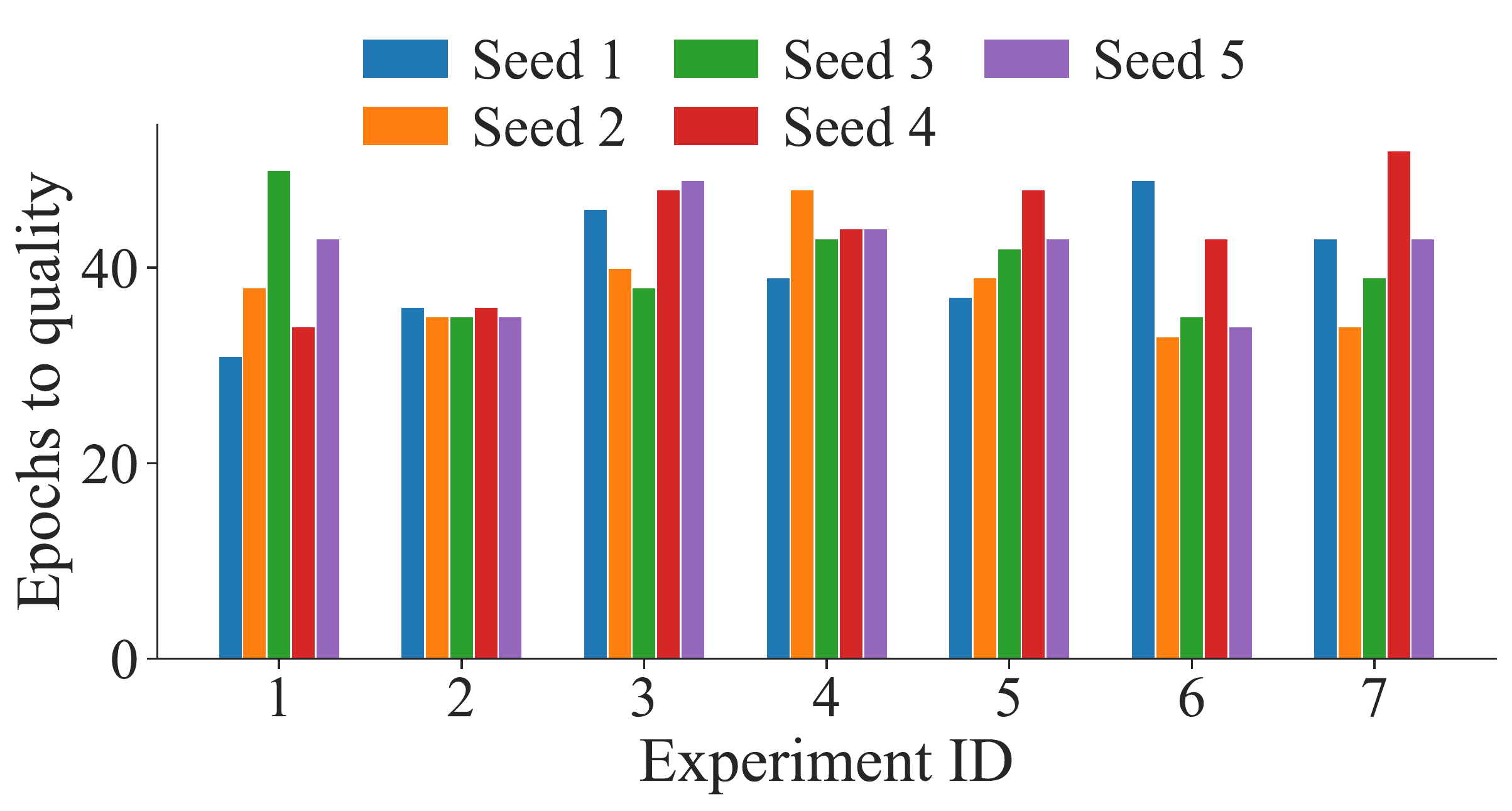}
  \caption{MiniGo.}
  \label{fig:epoch_variation_minigo}
  \end{subfigure}\vspace{-1mm}
  
  \caption{Training epochs to reach the target quality for the \mlperf v0.5 NCF (a) and MiniGo (b) benchmarks. Each experiment uses identical hyperparameters except for the random seed. For MiniGo, we observed considerable variability across runs even when fixing the random seed (same color). }
  \label{fig:epoch_variation}
\end{figure}

DNN training involves many stochastic influences that manifest in substantial run-to-run variation~\cite{choromanska2015loss, gori1992problem, auer1996exponentially, coleman2019analysis}. Different training sessions for the same model using the same hyperparameters can yield slightly different accuracies after a fixed number of epochs. Alternatively, different training sessions can take a different number of epochs to reach a given target accuracy. For example, Figure~\ref{fig:epoch_variation} shows the number of epochs needed to reach target accuracy for two \mlperf v0.5 benchmarks using reference implementations and default batch sizes. Several factors contribute to this variation, such as application behavior (e.g., random weight initialization and random data traversal) and system characteristics (e.g., profile-driven algorithm selection and the non-commutative nature of floating-point addition). Large distributed-training tasks can involve asynchronous updates, altering the gradient-accumulation order. These variations make it hard to reliably compare system performance.

\subsubsection{Diverse Software}
\label{sec:diverse_challenge}
Multiple ML software frameworks have emerged, each of which executes similar but distinct computations owing to various implementations and constraints~\cite{abadi2016tensorflow, paszke2017pytorch, chen2015mxnet, jia2014caffe}.
Software frameworks and the underlying math libraries employ different algorithms to implement the same operation. For example, convolutional and fully connected layers---two compute-intensive operators prevalent in modern DNN models---typically use cache blocking to exploit processor memory hierarchies. Different block sizes and processing orders (which optimize for different hardware), although algebraically equivalent, yield slightly divergent results.
In addition, operators can execute using various algorithms. For example, convolution layers can be executed using a variety of algorithms, including GEMM-based and transform-based (e.g., FFT or Winograd) variants.
In fact, the cuDNN v7.6 library provides roughly 10 algorithms for the forward pass of a convolutional layer,~\footnote{Source: cuDNN
(\url{https://docs.nvidia.com/deeplearning/sdk/cudnn-developer-guide}).} some of which vary in tiling or blocking choices depending on the hardware.
Although mathematically equivalent, different implementations will produce different numerical results, as floating-point representations have finite precision. 

Additionally, frameworks occasionally implement the same function in mathematically different ways. For example, modern training frameworks implement stochastic gradient descent with momentum in two ways:
\begin{equation}
\begin{split}
\text{momentum} & = \alpha \cdot \text{momentum} + \eta \cdot \frac{\partial L}{\partial w} \\
w& =w-\text{momentum}
\end{split}
\end{equation}
\begin{equation}
\begin{split}
\text{momentum} &= \alpha \cdot \text{momentum} + \frac{\partial L}{\partial w} \\
w &= w - \eta \cdot \text{momentum}
\end{split}
\end{equation}

The Caffe framework~\cite{jia2014caffe} implements the first approach, whereas PyTorch~\cite{paszke2017pytorch} and TensorFlow~\cite{abadi2016tensorflow} implement the second. These approaches differ mathematically if the learning rate $\eta$ changes during training---a common technique. Although this difference is tiny in many cases, it can hinder training convergence for larger minibatches.

Variations also arise owing to the frameworks' programming interface. For example, PyTorch and TensorFlow interpret asymmetric padding differently, complicating the task of porting model weights between them. Data-augmentation pipelines across frameworks can also apply image augmentations (e.g., crop, zoom, and rotation) in different orders.

Although ONNX~\cite{bai2019}, TVM~\cite{chen2018tvm}, and similar emerging tools enable interoperability of model architectures across frameworks, their support remains limited. Moreover, ML systems involve a range of optimizations that extend beyond the model architecture, such as preprocessing, precision, and communication methods. Benchmarks must accommodate the wide diversity of deployed systems despite this lack of a standard way to specify every training aspect. 

\subsection{Prior Work}
\label{sec:prior_work}

Prior ML benchmarks vary in granularity and scope. Microbenchmarks such as DeepBench~\cite{baidu2017deepbench} measure kernel-level operations that appear in commonly deployed models. Benchmarking such low-level operations fails to address the challenges associated with numerical precision, hyperparameter choices, and system scale, which we  described in the previous section. Furthermore, it neither captures the end-to-end application, nor accounts for memory- and cache-hierarchy effects across layers and operations, nor measures the data preprocessing that deep learning commonly employs.

Several benchmarks are defined at the granularity of entire DNN models. Fathom and Google TF Benchmarks~\cite{adolf2016fathom, google2017benchmarks} provide a reference suite of DNN models that span a wide application space, but they specifically measure model throughput and fail to account for accuracy. Similarly, TBD (Training Benchmarks for DNNs)~\cite{zhu2018tbd} profiles training on GPUs (but not other architectures) across diverse workloads, measuring characteristics such as memory and hardware utilization. Our benchmark builds on the diversity of applications in these projects while also capturing the quality and performance tradeoffs. 

DAWNBench~\cite{coleman2017dawnbench} was the first multi-entrant benchmark competition to use ``time to train'' (originally called time to accuracy) to measure the end-to-end performance of deep-learning systems; it allowed optimizations across model architectures, optimization procedures, software frameworks, and hardware platforms. Our benchmark follows a similar approach but handles more-diverse tasks (\S~\ref{sec:workloads}), and it uses important rules and mechanisms in the Closed division (\S~\ref{sec:division_reporting}) to enable fair comparisons of hardware and software systems.

Several other benchmarks are under development. AI Matrix measures workloads at different granularities (microbenchmarks, layer-wise benchmarks, end-to-end model benchmarks, and synthetic benchmarks)~\cite{aimatrix}. Deep500, although not a benchmark, provides a software framework for measuring DL-training performance~\cite{ben2019modular}.

\section{\mlperf Training Benchmark}
\label{sec:benchmark}
We now present the MLPerf Training benchmark, detailing the workloads (\S~\ref{sec:workloads}), timing rules (\S~\ref{sec:time_to_train}), quality-threshold choices (\S~\ref{sec:quality_thresholds}), and reference implementations and hyperparameters (\S~\ref{sec:references_hyperparameters}).

\subsection{Benchmark Suite}
\label{sec:workloads}
To create a fair and useful benchmark suite for modern ML workloads, we curated a representative set of tasks from several major ML areas, including vision, language, recommendation, and reinforcement learning. Our selection of benchmarks was primarily based on commercial and research relevance, representing diverse compute motifs. To establish relevance, we relied on feedback from the tens of commercial and academic organizations that support \mlperf. To keep the suite affordable, we selected a compact but representative set of seven benchmarks, which we describe below and summarize in Table~\ref{table:benchmarks}. Although these benchmarks already cover a wide range of research and industrial tasks, we are continuously exploring additional ones to keep the suite relevant to the ML community (\S~\ref{sec:discussion}).

\begin{table*}[t!]
\caption{\mlperf Training v0.5 benchmarks.}
\label{table:benchmarks}
\vskip 0.15in
\begin{center}
\begin{small}

\begin{tabular}{cccc}
\toprule

\bfseries Benchmark & \bfseries Data set & \bfseries Model & \bfseries Quality Threshold \\
\toprule
Image classification                                         & \makecell{ImageNet \\ \cite{deng2009imagenet}} & \makecell{ResNet-50 v1.5 \\ \cite{mlperf2017resnet}}                         & 74.9\% Top-1 accuracy               \\ \midrule
\makecell{Object detection \\ (lightweight)}                              & \makecell{COCO 2017 \\ \cite{lin2014microsoft}}& \makecell{SSD-ResNet-34 \\ \cite{liu2016ssd}}                          & 21.2 mAP                            \\ \midrule
\makecell{Instance segmentation and \\ object detection (heavyweight)}    & \makecell{COCO 2017 \\ \cite{lin2014microsoft}}& \makecell{Mask R-CNN \\ \cite{he2017mask}}           & \makecell{37.7 Box min AP, \\ 33.9 Mask min AP} \\ \midrule
\makecell{Translation \\ (recurrent)}                                      & \makecell{WMT16 EN-DE \\ \cite{wmt2016}}                      & \makecell{GNMT \\ \cite{wu2016google}}               & 21.8 Sacre BLEU                     \\ \midrule
\makecell{Translation \\ (nonrecurrent)}                                  & \makecell{WMT17 EN-DE \\ \cite{wmt2017}}       & \makecell{Transformer \\ \cite{vaswani2017attention}}& 25.0 BLEU                           \\ \midrule
Recommendation                                               & \makecell{MovieLens-20M \\ \cite{grouplens_2016}}                   & \makecell{NCF \\ \cite{he2017neural}}                                    & 0.635 HR@10                         \\ \midrule
Reinforcement learning                                       & \makecell{Go \\ (9x9 Board)}                   & \makecell{MiniGo \\ \cite{mlperf2017minigo}}                                 & 40.0\% Professional move prediction          \\
\bottomrule
\end{tabular}
\end{small}
\end{center}
\vskip -0.1in
\end{table*}

\subsubsection{Image Classification}
\label{sec:image_classification_workload}
Image classification is the most common task for evaluating ML-system performance~\cite{coleman2017dawnbench, adolf2016fathom, zhu2018tbd, goyal2017accurate, jia2018highly, mikami2018massively, ying2018image, google2017benchmarks, narayanan2019pipedream}. A classifier selects a class that best describes the contents of a given image. Classification model architectures also serve as feature extractors for many other computer-vision workloads, including  object detection, captioning, and style transfer. We use the ILSVRC 2012 ImageNet classification data set, consisting of 1.28 million training images and 50,000 validation images~\cite{deng2009imagenet}. Our model-quality metric is the Top-1 accuracy on the validation set. 

\minihead{ResNet-50} is a residual network~\cite{he2016deep, he2016identity}; such networks and their derivatives remain the state of the art in image classification, and system studies commonly use them~\cite{goyal2017accurate, jia2018highly, mikami2018massively, ying2018image, sun2019optimizing}. Several slightly different ResNet-50 implementations appear in training-framework repositories, preventing comparison of earlier system-performance claims because of model differences. To ensure meaningful system comparison, \mlperf uses the ResNet-50 v1.5 model, which performs addition after batch normalization, omits $1\times1$ convolution from the skip connection of the first residual block, and applies downsampling by the $3\times3$ convolutions. \mlperf also specifies the appropriate parameter initialization, optimizer schedule, and data augmentation.  

\subsubsection{Object Detection and Segmentation}
\label{sec:detection_segmentation_workload}
Object detection and segmentation are crucial components of many industrial systems for robotics, autonomous driving, video analytics, and social networks. Object detection is a regression task as opposed to a classification task: it returns bounding-box coordinates for objects in a given image. Segmentation assigns an object class to each input-image pixel. Although pretrained image-classification models commonly serve as the backbone (feature extractor) for DNN object detectors and segmenters, these DNN tasks differ from image classification in their compute characteristics. Examples include additional layer types (upscaling, ROIalign, NMS, and sorting); moreover, the inputs have greater resolution. \mlperf uses the 2017 COCO data set~\cite{lin2014microsoft} consisting of 118,000 training images and 5,000 validation images. Model-quality measurement uses mAP for both detection and segmentation.

\minihead{Mask R-CNN}~\cite{he2017mask} is a popular object-detection and instance-segmentation model for images. It has two stages: the first proposes regions of interest, and the second processes them to compute bounding boxes and segmentation masks. Mask R-CNN provides high-accuracy results for these tasks, but at the cost of higher latency as well as greater compute and memory requirements. The benchmark training uses images resized to 800 pixels on the shorter side and employs ResNet-50 as the backbone.

\minihead{Single Shot Detection (SSD)}~\cite{liu2016ssd} serves in real-time applications that require low-latency solutions. These applications include autonomous driving, robotics, and video analytics. Compared with Mask R-CNN~\cite{Huang2016} and other two-stage solutions, SSD trades speed for accuracy. Instead of full images, training uses $300\times300$ crops. We chose a ResNet-34 backbone to represent current real-time applications. ResNet-34 has a different residual-block structure than ResNet-50, increasing the diversity of computational motifs that \mlperf covers.

\subsubsection{Translation}
\label{sec:translation_workload}
Neural machine translation converts a sequence of words from the source language to a target language; many industrial applications employ this technology. As is common in translation research, we use the WMT English-to-German (EN-DE) data set~\cite{wmt2017}, which contains about 4.5 million sentence pairs. Our model-quality metric is the Bilingual Evaluation Understudy Score (Bleu) score on the Newstest2014 test set. We include two translation benchmarks to account for the two model architectures that translation and other sequence-data tasks often employ.

\minihead{Transformer}~\cite{vaswani2017attention} is an attention-based model that achieves state-of-the-art language-translation quality. It consists of an encoder and decoder, each being a stack of six blocks. Every block comprises a multihead attention layer and point-wise fully connected layers.

\minihead{GNMT}~\cite{wu2016google} is a recurrent neural network (RNN) for language translation. Even though it achieves lower accuracy than Transformer on the WMT English-to-German data set, it appears in the suite to represent RNN applications. These applications span numerous tasks, but language-translation data sets and publications are more common, enabling clearer system comparison. GNMT is the suite's only RNN. It consists of an eight-layer encoder and an eight-layer decoder, each using 1,024 LSTM cells with skip connections.

\subsubsection{Reinforcement Learning}
\label{sec:rl_workload}
Reinforcement learning (RL) is responsible for the recent dramatic increase in compute demand~\cite{amodei2018ai}, and it serves in control systems. RL algorithms can train agents (which includes neural networks) that rival humans at video games, go, and chess---major milestones in machine learning~\cite{silver2018general, mnih2013playing, chan_2018}. RL has a different computational profile than the other ML benchmarks: it generates training data through exploration instead of relying on a predetermined data set.

\minihead{MiniGo}~\cite{mlperf2017minigo}, inspired by AlphaGo~\cite{silver2016mastering, silver2017mastering, silver2018general}, trains a single model that represents both value and policy functions for a $9\times9$ game board. Training uses self-play (simulated games) between agents to generate data; rather than using a simulator, it performs many forward passes through the model to generate actions. We chose MiniGo to keep \mlperf more ML oriented, since many other RL problems employ simulators (physics, video-game environments, etc.) to generate data, spending most of their time in computations unrelated to ML. To measure quality, we calculate the percentage of predicted moves that match human reference games.

\subsubsection{Recommendation}
\label{sec:recommendation_workload}
Recommendation systems are a major commercial workload for Internet companies~\cite{naumov2019deep, zhou2018deep, cheng2016wide}. These workloads are characterized by large embedding tables followed by linear layers.

\minihead{Neural collaborative filtering (NCF)}~\cite{he2017neural} was our choice for the benchmark. It is trained to predict user-item interactions. More so than for other tasks, this recommender's compute characteristics depend on the data set. For example, the data set defines the embedding-table size as well as the memory-access patterns. Thus, a representative data set is crucial to a representative benchmark. Unfortunately, however, public data sets tend to be orders of magnitude smaller than industrial data sets. Although \mlperf v0.5 adopted the MovieLens-20M data set~\cite{grouplens_2016} for its NCF benchmark, v0.7 will employ a synthetically generated data set and benchmark while retaining the characteristics of the original data~\cite{belletti2019scalable}

\subsection{Time-to-Train Performance Metric}
\label{sec:time_to_train}
To address the ML-benchmarking challenges of system optimization and scale that we outlined in \S~\ref{sec:optimization_challenge} and \S~\ref{sec:scale_challenge}, \mlperf's performance metric is the time to train to a defined quality target. It incorporates both system speed and accuracy and is most relevant to ML practitioners. As an end-to-end metric, it also captures the auxiliary operations necessary for training such models, including data-pipeline and accuracy calculations. The metric's generality enables application to reinforcement learning, unsupervised learning, generative adversarial networks, and other training schemes. Time to train overcomes the challenges in \S~\ref{sec:optimization_challenge} and \S~\ref{sec:scale_challenge} by preventing submissions from using quality-reducing optimizations while still allowing for extensive system-scale and software-environment flexibility.

\subsubsection{Timing Rules}
\label{sec:timing_rules}
We chose the timing requirements to ensure fair system comparisons and to represent various training use cases. Timing begins when the system touches any training or validation data, and it stops when the system achieves the defined quality target on the validation data set. 

We exclude from timing several components that can carry substantial overhead and that are unrepresentative of real-world differences.

\minihead{System initialization.} Initialization, especially at large scales, varies on the basis of cluster-administrator choices and system-queue load. For example, it may involve running diagnostics on each node before starting the training job. Such overheads are unindicative of a system's training capability, so we exclude them from timing.

\minihead{Model creation and initialization.} Some frameworks can compile the model graph to optimize subsequent execution. This compilation time is insignificant for the longer training sessions when using industry-scale data sets. \mlperf, however, uses public data sets that are usually much smaller than industry ones. Therefore, large distributed systems can train some \mlperf benchmarks in minutes, making compilation times a substantial portion of the total time. To make benchmarks representative of training on the largest industrial data sets, we allow exclusion of up to 20 minutes of model-creation time. This limit ensures that \mlperf captures smaller training jobs, and it discourages submissions with compilation approaches that are too computationally and operationally expensive to use in practice.

\minihead{Data reformatting.} The raw input data commonly undergoes reformatting once and then serves in many subsequent training sessions. Reformatting examples include changing image-file formats and creating a database (e.g., LMDB, TFRecords, or RecordIO) for more-efficient access. Because these operations execute once for many training sessions, \mlperf timing excludes reformatting. But it prohibits any data processing or augmentation that occurs in training from moving to the reformatting stage (e.g., it prevents different crops of each image from being created and saved before the timed training stage).

\subsubsection{Number of Timing Runs}
\label{sec:number_of_timing_samples}
To address the stochastic nature and resulting run-to-run variance of modern deep-learning methods described in \S~\ref{sec:variation_challenge}, \mlperf requires that submissions provide several runs of each benchmark to stabilize timing. We determined the number of runs, which varies among benchmarks, by studying the behavior of reference implementations. Vision tasks require 5 runs to ensure 90\% of entries from the same system are within 5\%; all other tasks require 10 runs to ensure 90\% of entries from the same system are within 10\%. \mlperf drops the fastest and slowest times, reporting the arithmetic mean of the remaining runs as the result.

\subsection{Choice of Quality Thresholds}
\label{sec:quality_thresholds}
\begin{figure}[t!]
  \centering
  \includegraphics[width=0.85\columnwidth]{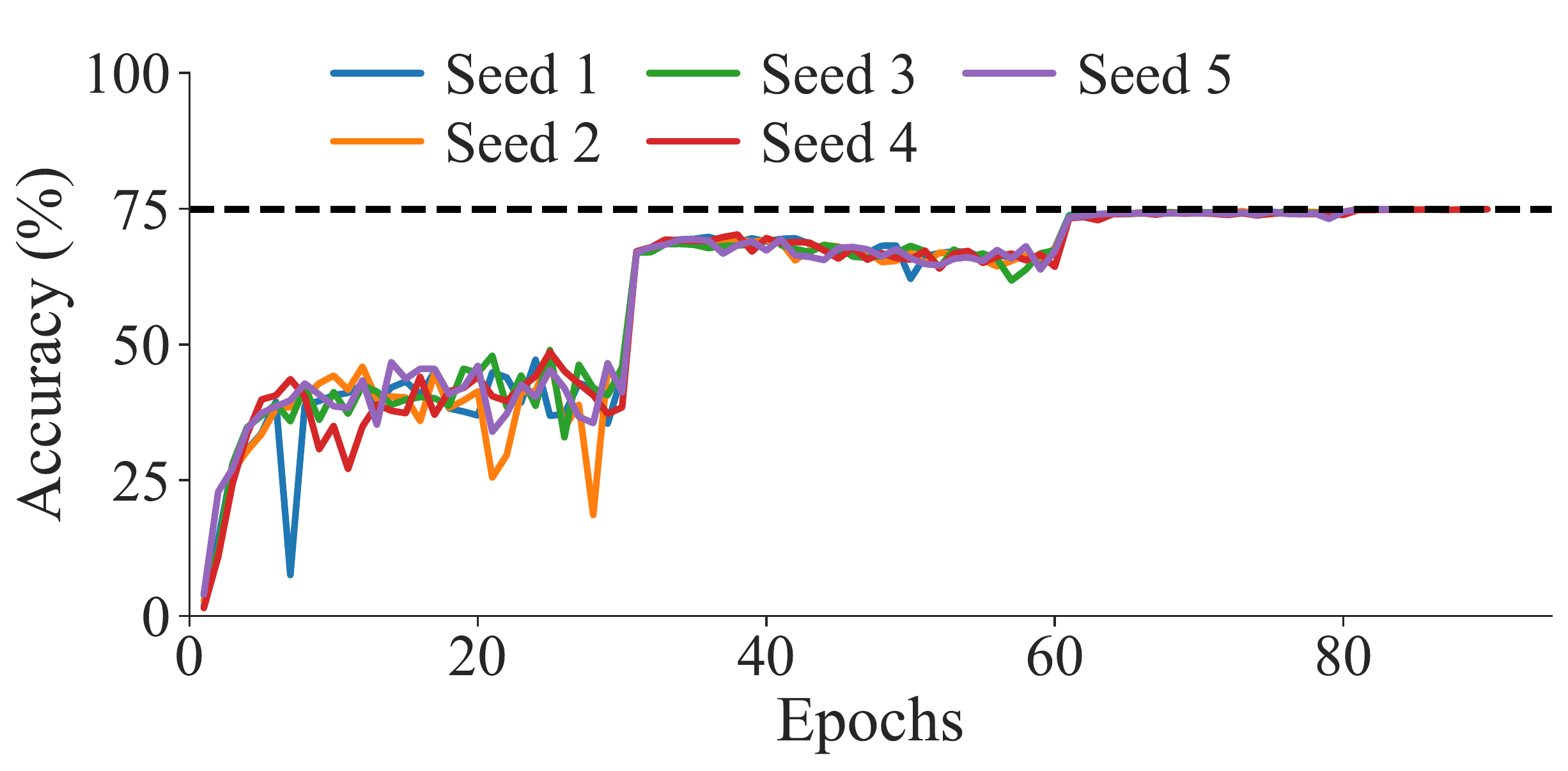}
  \caption{Top-1 accuracy of \mlperf v0.5 ResNet-50 benchmark over 100 epochs for five runs (denoted by color) with identical hyperparameters but different random seeds. The dashed line indicates the quality target of 74.9\% Top-1 accuracy. The early training phase exhibits much more variability than later phases.}
  \label{fig:different_resnet_seeds}
\end{figure}

For each benchmark, we chose quality metrics near the state of the art for the corresponding model and data set (Table~\ref{table:benchmarks}), basing our choice on experiments with the reference implementations. Some of these thresholds are slightly lower than results in the literature, enabling us to benchmark across software frameworks and to ensure that training sessions consistently achieve the quality metric. Although selecting a lower threshold that is achievable earlier in a training session reduces submission resources, we chose higher thresholds that require longer training sessions for two reasons: First, we must prevent optimizations from adversely affecting the final results (challenges described in \S~\ref{sec:optimization_challenge} and \S~\ref{sec:scale_challenge}). Second, we must minimize run-to-run variation, which tends to be much higher early in training. For example, Figure~\ref{fig:different_resnet_seeds} shows accuracy for five training sessions of \mlperf v0.5's ResNet-50 v1.5 reference implementation, where the first 30 epochs exhibit considerably more noise.

\subsection{References and Hyperparameters}
\label{sec:references_hyperparameters}
\mlperf provides a reference implementation for each benchmark, using either the PyTorch or TensorFlow framework. References also include scripts or directions to download and preprocess public data sets. References are not optimized for performance (meaning they should not be used for performance assessment or comparison), as their main purpose is to define a concrete implementation of a benchmark model and training procedure. All submitters must follow these references---they may reimplement a benchmark in their framework of choice as long as the DNN model and training operations are mathematically equivalent to the reference. Furthermore, \mlperf uses reference implementations to establish the required quality thresholds.

\begin{table}[t!]
\caption{\mlperf modifiable hyperparameters.}
\label{table:hyperparameters}
\vskip 0.15in
\begin{center}
\begin{small}

\begin{tabular}{L{2.5cm}L{4.5cm}}
\toprule

\bfseries Model & \bfseries Modifiable Hyperparmeters \\
\toprule
\makecell{All that use SGD} & Batch size, Learning-rate schedule parameters\\ \midrule
\makecell{ResNet-50 v1.5} &  \\ \midrule
\makecell{SSD-ResNet-34} & Maximum samples per training patch \\ \midrule
\makecell{Mask R-CNN} & Number of image candidates \\ \midrule
\makecell{GNMT} & Learning-rate decay function, Learning rate, Decay start, Decay interval, Warmup function, Warmup steps \\ \midrule
\makecell{Transformer} & Optimizer: Adam \cite{kingma2014adam} or Lazy Adam, Learning rate, Warmup steps  \\ \midrule
\makecell{NCF} & Optimizer: Adam or Lazy Adam, Learning rate, $\beta1$, $\beta2$ \\ \midrule
\makecell{Go (9x9 board)} & \\ 
\bottomrule
\end{tabular}
\end{small}
\end{center}
\vskip -0.1in
\end{table}

\mlperf rules specify the modifiable hyperparameters (Table~\ref{table:hyperparameters}) as well as restrictions on their modification. These restrictions are intended to balance the need to tune for different systems with limiting the size of the hyperparamter search space to be fair to submitters with smaller compute resources. For example, to accommodate a wide range of training-system scales, submissions must be able to adjust the minibatch size used by SGD in order to showcase maximum system efficiency (this approach is similar in concept to the Top500 LINPACK benchmark, which allows systems to choose the problem size). To ensure that training still converges to the required threshold, other hyperparameters---such as the learning rate schedule---may need adjustment to match. For example, a common ResNet training practice is to to increase the learning rate linearly with the minibatch size~\cite{goyal2017accurate}. Although these hyperparameter searches are a common ML task, \mlperf's focus is on system optimization rather than hyperparameter exploration and we do not want to penalize submitters who are unable to do extensive searches. Therefore we restrict hyperparamter tuning to subset of all possible parameters and values.

Further, we allow ``hyperparameter borrowing'' during the post-submission review process in which one submitter may adopt another submitter's hyperparamters for a specific benchmark and resubmit their result (with no other hardware or software changes allowed). In the first two rounds, hyperparameter borrowing was used successfully to improve several submissions indicating hyperparamters are somewhat portable. Typically borrowing occured across systems of similiar scale, but did result in convergence across different numerics (FP16, bfloat16, and FP32), architectures (CPU, GPU, and TPU), and software implementations (TF, cuDNN, and MKL-DNN). \mlperf working groups review the hyperparameter choices and requirements for each benchmark round to account for advances in training ML models at scale.

\section{Benchmarking Process}
\label{sec:process}
Next, we outline the process for submission and review (\S~\ref{sec:submission_and_review}) and for reporting results (\S~\ref{sec:results_reporting}) to account for innovative solutions, availability, and scale.
We have run two rounds of the \mlperf benchmark: v0.5 and v0.6. The time between rounds is about a few months, allowing us to update the suite after each one. Every round has a submission and review period followed by publication of results.

\subsection{Submission and Review}
\label{sec:submission_and_review}
An \mlperf submission consists of a system description, training-session log files, and all code and libraries required to reproduce the training sessions. All of this information is publicly available on the \mlperf GitHub site, along with the \mlperf results, allowing for reproducibility and enabling the community to improve the results in subsequent rounds. A system description includes both the hardware (number of nodes, processor and accelerator counts and types, storage per node, and network interconnect) and the software (operating system as well as libraries and their versions). A training-session log file contains a variety of structured information including time stamps for important workload stages, quality-metric evaluations at prescribed intervals, and hyperparameter choices. These logs are the foundation for analyzing results.

Before publishing results, submissions are peer-reviewed for compliance with \mlperf rules. Submitters receive notification of noncompliance, where applicable, and they may resubmit after addressing any such problems. Additionally, we permit some hyperparameter borrowing as described earlier during this period.

\subsection{Reporting Results}
\label{sec:results_reporting}
Each \mlperf submission has several labels: division (open or closed), category (available, preview, or research), and system type (on-Premises or cloud).

\subsubsection{Submission Divisions}
\label{sec:division_reporting}
\mlperf has two submission divisions: closed and open. Both require that submissions employ the same data set and quality metric as the corresponding reference implementation.

The \emph{closed} division is intended for direct system comparison, so it strives to ensure workload equivalence by requiring that submissions be equivalent to reference implementations. Equivalence includes mathematically identical model implementations, parameter initialization, optimizer and training schedules, and data processing and traversal. To ensure fairness, this division also restricts hyperparameter modification.

The \emph{open} division is intended to encourage innovative solutions of important practical problems and to encourage hardware/software co-design. It allows submissions to employ model architectures, optimization procedures, and data augmentations that differ from the reference implementations.

\subsubsection{System Categories}
\label{sec:category_reporting}
To allow for a broad range of research and industry systems, we defined three submission categories: available, preview, and research. These categories encourage novel techniques and systems (e.g., from academic researchers), but they also distinguish between shipping products and proof-of-concept or early engineering samples.

The \emph{available} category imposes requirements on both hardware and software availability. Hardware must be either available for third-party rental on a cloud service or, in the case of on-premises equipment, available for purchase. Supply and lead times for renting or purchasing should befit the system scale and company size. To ensure that benchmark submissions are widely consumable and to discourage benchmark-specific engineering, we also require that software in this category be versioned and supported for general use.

\emph{Preview} systems contain components that meet the available-category criteria within 60 days of the submission date or by the next submission cycle, whichever is later. Any preview system must also be submitted to the available category by that time. 

\emph{Research} submissions contain components unintended for production. An example is an academic-research prototype designed as a proof of concept rather than a robust product. This category also includes systems that are built from production hardware and software but are larger in scale than available-category configurations.

\subsubsection{Reporting Scale}
\label{scale_reporting}
Modern ML training spans multiple orders of magnitude in system power draw and cost. Therefore, comparisons are more useful if the reported performance includes the scale. A common scale metric, such as cost or power, is not definable across a wide range of systems (cloud, on-premises, and preproduction), so it requires differentiation by system type.

In the first two \mlperf rounds, we included the system configuration (number of processors and/or accelerators) alongside the performance scores. For on-premises examples, future versions will include a power-measurement specification. For cloud systems, we derived a ``cloud-scale'' metric from the number of host processors, amount of host memory, and number and type of accelerators. We empirically verified that cloud scale correlates closely with cost across three major cloud providers. Reporting of these scale metrics was optional in \mlperf v0.5 and v0.6. 

\subsubsection{Reporting Scores}
\label{summary_reporting}
An \mlperf results report provides the time to train for each benchmark. Although a single summary score that spans the entire suite may be desirable for system comparisons, it is unsuited to \mlperf for two main reasons. First, a summary score implies some weighting of individual benchmark scores. Given the diversity of system users and the wide range of applications that \mlperf covers, no weighting scheme is universally representative. Second, a summary score becomes less meaningful if a submitter declines to report results on all benchmarks. Submitters can have multiple reasons for omitting some benchmarks---not all are practical at every system scale (for example, some models are untrainable at the minibatch sizes that the largest systems require for data-parallel training). Additionally, some processors may target only certain applications.

\section{Results}

\begin{figure}[t!]
  \centering
  \begin{subfigure}[c]{0.85\columnwidth}
    \includegraphics[width=1.0\columnwidth]{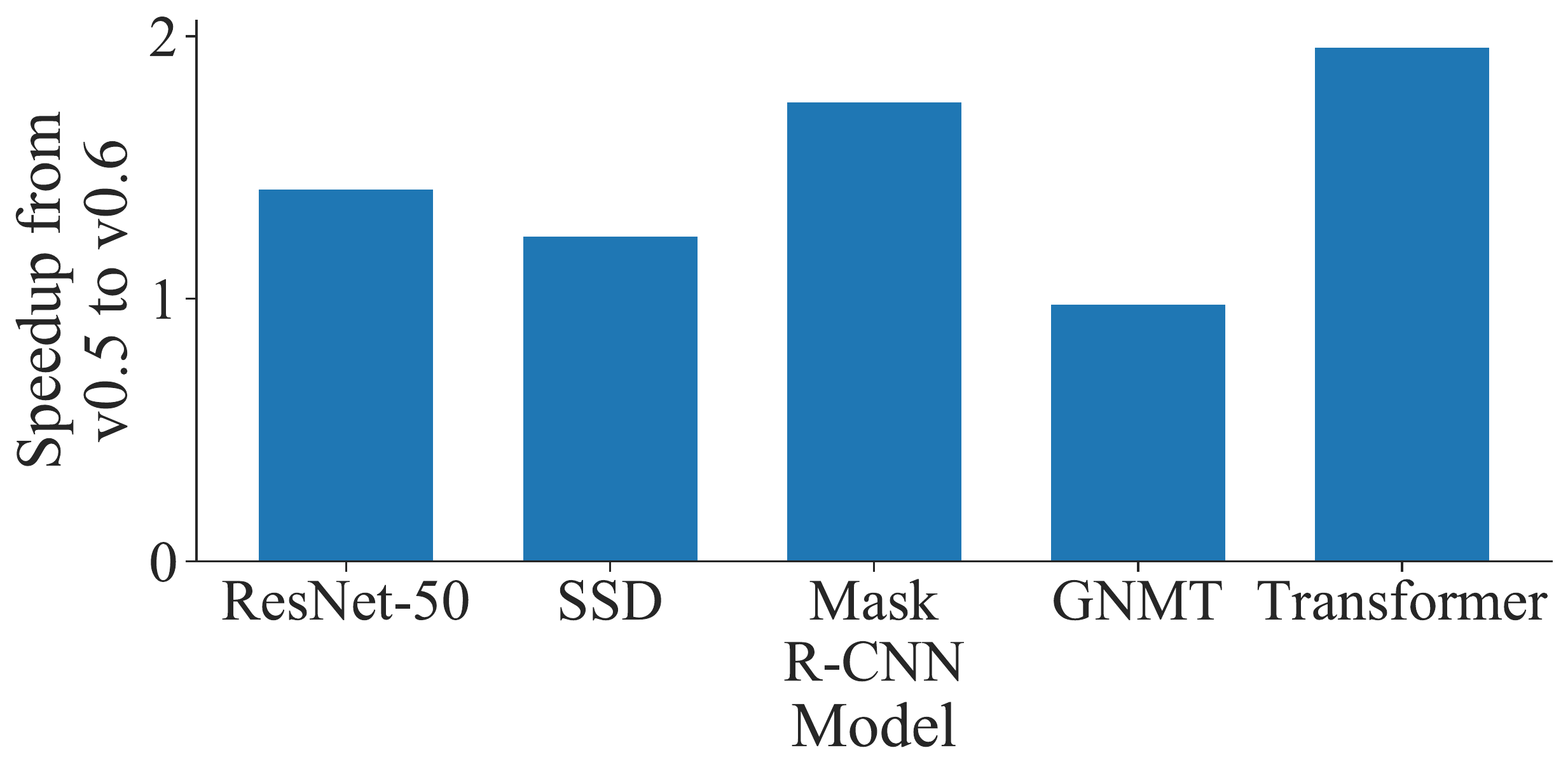}
    \caption{Speedup.}
  \end{subfigure}
    
  \begin{subfigure}[c]{\columnwidth}
        \footnotesize
        \centering
        \begin{tabular}{lrrr}
        \toprule
        Model  &  Metric & v0.5 & v0.6      \\
        \toprule
        ResNet-50 & Top-1 accuracy &  74.9 & 75.9 \\
        SSD & mAP & 21.2 & 23 \\
        Mask R-CNN & Box / Mask min AP & 37.7 / 39.9 & Same \\
        GNMT & Sacre BLEU & 21.8 & 24 \\
        Transformer & BLEU & 25 & Same \\
 
        \bottomrule
        \end{tabular}
\caption{Quality targets.}
  \end{subfigure}
  \caption{Speedup in the fastest 16-chip entry from \mlperf version v0.5 to v0.6 for various benchmarks common to both (Figure 3a), along with quality-target increases (Figure 3b).}
  \label{fig:speedup_between_mlperf_versions}
\end{figure}

\begin{figure}[t!]
  \centering
  \includegraphics[width=0.85\columnwidth]{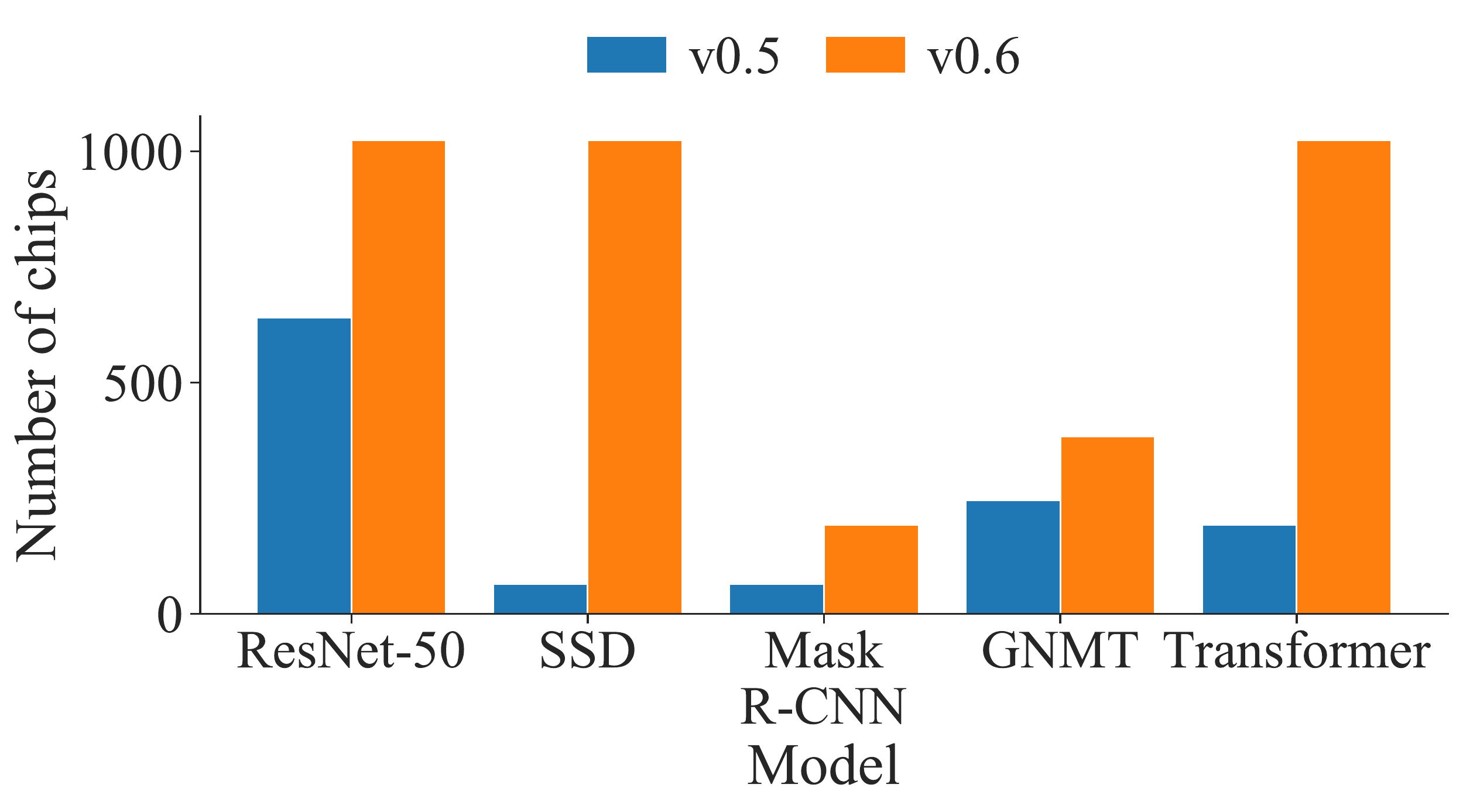}
  \caption{Number of chips necessary to produce the fastest time to solution for \mlperf versions v0.5 to v0.6. This number increased by as much as $5.5\times$.}
  \label{fig:chip_count_between_mlperf_versions}
\end{figure}

\label{sec:results}
\mlperf, like all benchmarks, aims to to encourage innovation through constructive competition; we measure progress by comparing results across submission rounds. We have conducted two MLPerf Training rounds thus far: v0.5 and v0.6. They were six months apart, and the underlying hardware systems were unchanged. The results that were either unmodified or underwent minor modifications between rounds show that \mlperf is driving rapid performance and scaling improvement in both the implementations and software stacks. Figure~\ref{fig:speedup_between_mlperf_versions} shows that between the two submission rounds, the best performance results for a 16-chip system increased by an average of $1.3\times$ despite the higher quality targets. Figure~\ref{fig:chip_count_between_mlperf_versions} reveals that the number of chips necessary to produce the best overall performance result increased by an average of $5.5\times$. Some of this improvement owes to better benchmark implementations and some to rule changes, such as allowing the LARS~\cite{you2017large} optimizer for large ResNet batches. But we believe submitters incorporated much of the performance and scaling improvements into the underlying software infrastructure and passed them on to users. We expect \mlperf to drive similar improvements through focused hardware innovation.

\section{Conclusions}
\label{sec:discussion}
\mlperf Training is a suite of ML benchmarks that represent both industrial and academic use cases. In addition to being the only widely used ML-training benchmark suite boasting such coverage, it has made the following contributions:
\begin{itemize}[topsep=0pt]
\item Precise definition of model architectures and training procedures for each benchmark. This feature enables system comparisons for equivalent workloads, whereas previous results often involved substantially different variants of a given model (for example, ResNet-50 has at least five variants).
\item Reference implementations and rule definitions to address the challenges unique to benchmarking ML training. These challenges include the stochastic nature of training processes, the necessity of training to completion to determine the quality impact of performance optimizations, and the need for workload variation at different system scales (\S~\ref{sec:challenges}).
\end{itemize}

Although \mlperf focuses on relative system performance, as the online results demonstrate, it also offers general lessons about ML and benchmarking: 

\begin{itemize}[topsep=0pt]
\item Realistic data-set size is critical to ensuring realistic memory-system behavior---for example, the initial NCF data set was too small and could reside entirely in memory. Furthermore, when benchmarking data sets that are smaller than industrial scale, training time should exclude the startup time, which would be proportionally less in actual use. 
\item Small hyperparameter changes can produce considerable performance changes. But, based on our experience with hyperparameter borrowing, hyperparameters are relatively portable at similiar system scales, even across architectures, numerics, or software stacks.
\item Frameworks exhibit subtle optimizer-algorithm variations that affect convergence.
\end{itemize}

ML is an evolving field, however, and we have much more to learn. To keep pace, \mlperf establishes a process to maintain and update the suite. For example, \mlperf v0.6 includes several updates: the ResNet-50 benchmark added LARS~\cite{you2017large}, GNMT's model architecture improved to increase translation quality, and the MiniGo reference switched from Python to C++ to increase performance. The \mlperf organization welcomes input and contributions:~\url{https://mlperf.org/get-involved}

\section*{Acknowledgements}
In this section, we acknowledge all those who helped produce the first set of results or supported the overall
benchmark development.

\minihead{Intel:}
Cong Xu,
Deng Xu,
Feng Tian,
Haihao Shen,
Mingxiao Huang,
Rachita Prem Seelin,
Teng Lu,
Xin Qiu,
and Zhongyuan Wu.

\minihead{Facebook:}
Maxim Naumov,
Dheevatsa Mudigere,
Mustafa Ozdal,
Misha Smelyanskiy,
Joe Spisak,
Sy Choudhury,
and Brian Gamidos.

\minihead{Stanford:}
Work at Stanford received support in part from affiliate members and other Stanford DAWN project participants---Ant Financial, Facebook, Google, Infosys, NEC, and VMware---as well as Toyota Research Institute, Northrop Grumman, Cisco, SAP, NSF CAREER grant CNS-1651570, and  NSF Graduate Research Fellowship grant DGE-1656518. Any opinions, findings, conclusions, or recommendations expressed in this material are those of the authors and do not necessarily reflect the views of the NSF.

\minihead{Harvard:}
Work at Harvard received partial support from the Applications Driving Architectures (ADA) Research Center, a JUMP Center cosponsored by the SRC and DARPA, NSF CCF\#1704834, and Intel Corporation.
We would also like to thank Brandon Reagen.

\minihead{University of Toronto:}
Work at the University of Toronto received partial support from an NSERC Discovery grant, the Canada Foundation for Innovation JELF grant, the Connaught Fund, and Huawei grants.

\bibliography{sysml2019}
\bibliographystyle{mlsys2020}

\clearpage
\appendix
\section{Artifact Appendix}

\subsection{Abstract}

This artifact description contains information about the complete workflow to reproduce Nvidia's v0.5 image classification submissions to MLPerf. We describe how to run this submission on a single-node DGX-1 system. More details for DGX-2 and multi-node systems are provided in the official MLPerf results repositories:

\begin{itemize}
    \item \href{https://github.com/mlperf/training_results_v0.5/tree/master/v0.5.0/nvidia/submission/code/image_classification/mxnet}{Nvidia's v0.5 ResNet-50 submissions}
\end{itemize}

Results from other tasks and submitters are also available:

\begin{itemize}
    \item \href{https://github.com/mlperf/training_results_v0.5}{MLPerf v0.5 training results}
    \item \href{https://github.com/mlperf/training_results_v0.6}{MLPerf v0.6 training results}
\end{itemize}

However, these results have not been independently verified for reproducibility. Please see the MLPerf website (\url{https://mlperf.org/}) for the most up-to-date information and feel free to report issues on Github.

\subsection{Artifact check-list (meta-information)}

{\small
\begin{itemize}
  \item {\bf Algorithm: } Image classification ResNet-50 CNN
  \item {\bf Program: } MLPerf (\url{https://mlperf.org/})
  \item {\bf Compilation: } \href{https://github.com/NVIDIA/nvidia-docker}{nvidia-docker}
  \item {\bf Model: } ResNet-50 v1.5~\footnote{\url{https://github.com/mlperf/training/tree/master/image_classification/tensorflow/official}}
  \item {\bf Data set: } ImageNet (\url{http://image-net.org/})
  \item {\bf Hardware: } NVIDIA DGX-1 or DGX-2
  \item {\bf Metrics: } Time-to-Train: minutes to reach accuracy threshold (74.9\% Top-1 for v0.5)
  \item {\bf Output: } MLPerf compliant log file with timestamps and evaluation accuracy. Execution ends once the accuracy threshold is reached.
  \item {\bf Experiments: } shell script included with the code (./run.sub)
  \item {\bf How much disk space required (approximately)?: } 300 GB
  \item {\bf How much time is needed to prepare workflow (approximately)?: } 2 hours
  \item {\bf How much time is needed to complete experiments (approximately)?: } 8 hours
  \item {\bf Publicly available: } Yes
  \item {\bf Code licenses: } Apache License 2.0
  \item {\bf Workflow framework used?: } MXNet
  \item {\bf Archived (provide DOI)?: } \\\url{http://doi.org/10.5281/zenodo.3610717}
\end{itemize}

\subsection{Description}

\subsubsection{How to access}

MLPerf v0.5 training results on Github: \\
\url{https://github.com/mlperf/training_results_v0.5}.





\subsection{Installation}

See the README.md for Nvidia's v0.5 ResNet-50 submission: \url{https://github.com/mlperf/training_results_v0.5/tree/master/v0.5.0/nvidia/submission/code/image_classification/mxnet/README.md}.


\subsection{Evaluation and expected result}

Time-to-Train: 134.6 minutes.


\end{document}